# Deep Learning Models Delineates Multiple Nuclear Phenotypes in H&E Stained Histology Sections

Mina Khoshdeli, *member, IEEE*, and Bahram Parvin, *Senior Member, IEEE*

*Abstract*—Nuclear segmentation is an important step for profiling aberrant regions of histology sections. However, segmentation is a complex problem as a result of variations in nuclear geometry (e.g., size, shape), nuclear type (e.g., epithelial, fibroblast), and nuclear phenotypes (e.g., vesicular, aneuploidy). The problem is further complicated as a result of variations in sample preparation. It is shown and validated that fusion of very deep convolutional networks overcomes (i) complexities associated with multiple nuclear phenotypes, and (ii) separation of overlapping nuclei. The fusion relies on integrating of networks that learn region- and boundary-based representations. The system has been validated on a diverse set of nuclear phenotypes that correspond to the breast and brain histology sections.

*Index Terms*—nuclear segmentation, deep convolutional neural networks, color decomposition, vesicular phenotype

## I. INTRODUCTION

NUCLEAR morphology is an important step in identifying aberrant phenotypes in hematoxylin and eosin (H&E) stained histology sections. However, to date, the problem of nuclear segmentation, for every type of nuclear phenotype remains unresolved. If nuclear segmentation is preformed robustly, then cell types and tissue architecture can be represented faithfully for normal and malignant phenotypes, and tumor subtypes and their microenvironments can be stratified across a large cohort. The main challenges originate from technical variations and biological heterogeneity in a large cohort. Technical variations refer to non-uniformity in sample preparations and fixation, and biological heterogeneity refers to the fact that no two histology sections are alike. In most cases, technical variations are also coupled with biological heterogeneity, which complicates the construction of a stable computational model for nuclear segmentation. The diversity of the nuclear phenotypes originates from many factors. For example, (i) cancer cells tend to be larger than normal cells, and if coupled with high chromatin content, they may indicate aneuploidy; (ii) nuclei may have a vesicular phenotype; (iii) nuclei may have high pleomorphism in tumor sections; (iv) cells may be going through apoptosis or necrosis; (v) cell cytoplasm may be lost as a result of clear cell carcinoma; and (vi) cellular phenotype may be altered as a result of macromolecules being secreted into the microenvironment.

B. Parvin and M. Khoshdeli are with the Department of Electrical and Biomedical Engineering, University of Nevada, Reno, NV, 80557 USA.

These issues suggest complexities that are associated with nuclear segmentation as one of the steps toward profiling of histology sections for diagnostic and discovery of new biomarkers. Because of the complexities associated with vesicular phenotypes, most of the previous segmentation literature has focused on nuclear phenotypes having high DNA contents. However, we show that simultaneous delineation of vesicular and other phenotypes are overcome with the deep learning models.

In recent years, convolutional neural networks (CNN)s have emerged as the most powerful technique for image classification [1, 2], and image segmentation [3-6]. CNNs can be continuously trained and improved as the number of annotated training samples increases. Furthermore, their architecture is modular, where each module can be trained for different image-based representation, and modules can be integrated to improve the outcome. For example, one CNN module can be trained for the boundary-based representation of the objects of interest, a second CNN module can be trained by the underlying spatial distribution of the object, and a third CNN module can be trained for the location of an object in terms of its bounding box. Collectively, integration of these three modules can improve the segmentation task.

Consequently, because of the diversity of the nuclear phenotypes and complexities associated with modeling the appearance of nuclear morphology, CNN has the potential to map and capture these diversities through generative models, i.e., training and automatic feature learning. We propose a model based on deep CNN for nuclear segmentation with the integration of boundary- and region-based information, i.e., a fusion of CNNs constructed by region-based and boundary-based segmentation. The region-based model includes a deep convolutional neural model, which labels foreground (nuclei) and background pixels. Although the region-based model could not distinguish touching or overlapping nuclei, the boundary-based model can produce boundary features or perceptual boundary features. Finally, the last step of the framework is to fuse the information from the two models, where fusion is performed by training another convolutional neural network. **Figure 1** illustrates the general framework of the proposed nuclear segmentation method. The main advantages of the proposed model are (i) enabling segmentation of nuclear phenotype with vesicular phenotype, (ii) high throughput processing that will be necessary for processing of whole slide images; and (iii) contributing to the separation of touching or overlapping nuclei by incorporating boundary- and region-based information. Organization of this paper is as follows: Section II reviews previous research. Section III describes the details of the proposed deep learning model. Section IV presents our experimental results. Lastly, Section V concludes the paper.

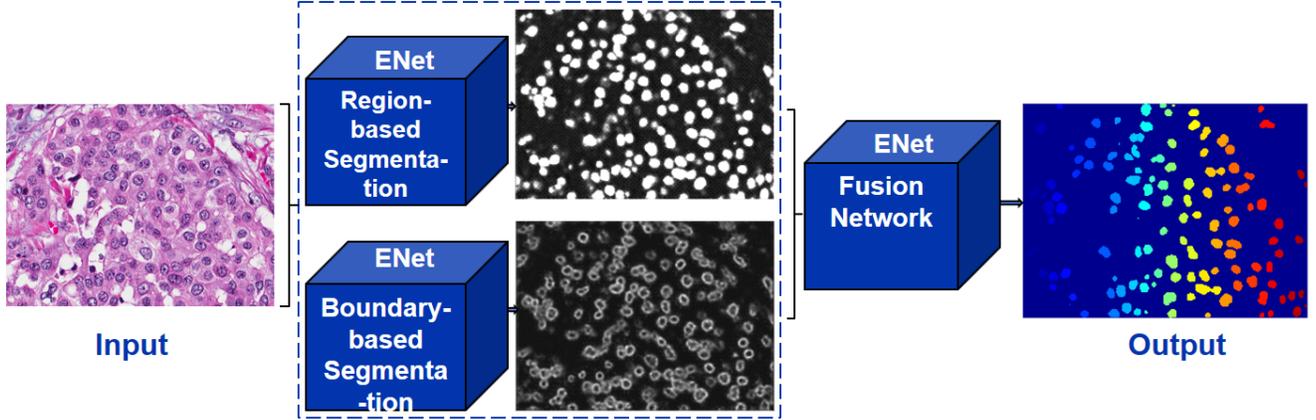

Fig. 1. Nuclear segmentation architecture includes three ENets. Two ENets are used for region-based and boundary-based segmentation. The outputs of these two networks are then fused through a third ENet.

## II. REVIEW OF PREVIOUS WORKS

There are two comprehensive review papers on the nuclear segmentation techniques [7, 8]; therefore, we limit ourselves to a summary here, which span from simple thresholding to the application of convolutional neural networks.

The most popular nuclear segmentation approaches include thresholding following morphological operations [9, 10], watershed [11], deformable models [12], and graph-based models [13, 14] or a combination of these methods. In[10], images are binarized, morphological operators are applied, and nuclear features have been computed to profile the tumor morphology. In [11], the watershed segmentation has been applied to the magnitude of the gradient image, where the initial seeds have been generated by morphological operations. This technique is very dependent on the initial seeds, and over-segmentation may occur due to non-uniform nuclear regions. In [12], an efficient active contour model was proposed; however, this technique would not work well for nuclei having a vesicular phenotype. Similar methods have also been proposed with multi-step graph cut formulation [14], but the key assumption remains nuclei with high chromatin content, In [13], Gaussian Mixture Model (GMM) models of nuclear phenotypes were constructed by interactively annotating nuclear regions. The GMM representation was based on the LoG response and the RGB values in the color space. Next, a multi-reference graph cut method was developed to binarize the image. Subsequently, each clump of overlapping nuclei was partitioned using geometric reasoning. Because this technique is model-based, intrinsic variations of colony and intensity are captured in GMM. A similar method was in proposed [15]; however, thresholding was based on training parameters of the model using support vector machine (SVM); however, graph cut has the additional advantage over SVM since it incorporates spatial consistency. In summary, except the last two methods, most classical techniques are procedural and model-free with a large number of free parameters. Model-based methods [13, 15] overcome complexities associated with the technical variations and biological heterogeneity to a certain extent with the net result being an improved performance profile.

There are wide applications of CNN for processing of stained histology sections that include (a) nuclear detection, (b) nuclear segmentation, and (c) gland segmentation.

(a) With respect to nuclear detection, three strategies are reviewed here. In [16], a spatially constrained CNN model has been trained for nuclear detection. The model has been spatially constrained by assigning a higher probability to the pixels that are closer to the centroids of nuclei. A similar approach has been proposed in [17], where a CNN model is trained to generate the positions of the nuclei and their corresponding confidence in a given patch. In [18], a CNN model has been trained with the feature-based representation of the original image based on the Laplacian of Gaussian (LoG) filter response. The advantage of the LoG filter is that it accentuates the blob-shape of nuclei and provides an approximate location of each nucleus. This model has been applied to detect various types of the nuclear phenotype.

(b) With respect to nuclear segmentation, CNN models have been trained for region-based segmentation, semantic-level feature extraction, and nuclear segmentation. In [19], an active contour model has been utilized for nuclear segmentation in H&E stained breast histology sections, and a CNN model has been trained to extract semantic-level features and to make an initial classification of the image into the low, intermediate, or high-grade tumor. Subsequently, the final classification is refined by integrating semantic-level (e.g., the ratio of nuclei belonging to different grades), colony organization level (e.g., the relationship of nuclei within and across colonies), and pixel-level (e.g., texture) features to train an SVM. However, the active contour model assumes that nuclei are well isolated and have high chromatin, but this is not the case in practice. In fact, for breast cancer, nuclear atypia is one of the visual representation for grading. In [20], a CNN-based model has been proposed for nuclei segmentation from H&E stained sections, where the CNN is trained to classify each pixel to be nuclei or non-nuclei.

In [21], a multiscale convolutional network has been proposed for the segmentation of cervical cytoplasm and nuclei. The multiscale CNN incorporates a pyramid image representation for initial pixel-based classification. Next graphcut is applied since CNN does not enforce spatial continuity. Finally, segmentation results are refined by morphological operators such as a marker-based watershed. In [22], Nuclear segmentation has been performed by converting the RGB image into gray scale, denoising the image, and applying the CNN to separate background and foreground. Finally, nuclear segmentation is refined by morphological operators. A similar approach has been proposed in [23], a CNN based model has been trained to provide the initial probability map for nuclear segmentation. Then, a deformable shape model has been applied to separate overlapping nuclei.

(c) With respect to the gland segmentation, a controlled study was performed to demonstrate that CNN outperforms engineered feature extraction coupled with SVM [24]. Another group showed an improved performance by fusing multiple CNNs that are trained with different segmentation objective in terms of regions, boundaries, and detection [25].

We propose a fusion model based on very deep CNN for nuclear segmentation from hematoxylin and eosin (H&E) stained histology sections. The proposed model is different from previous research in three ways. First, previous researchers have trained CNN models to infer a probability vector for each pixel from a sliding input patch. In contrast, our approach is not based on mapping the center pixel of a patch to a unique classification label. Second, proposed networks are very deep, consisting of 17 layers, and all convolutions are either 3-by-3 or one-dimensional. As a result, training and testing are expedited because of the low dimensionality of convolution operations. Third, different representations of the underlying spatial distributions are computed, where each representation has its own deep learning model. The representations incorporate both region- and boundary-based information.

The basic computational unit of the proposed learning model has an encoder-decoder architecture, leverages an extension of ENet [4], and has 17 layers for with small convolutional operations; thus, enabling efficient training and instance segmentation during testing. The ENet model also has the advantages of using various types of convolution operations that include regular, asymmetric, and dilated. The diversity of convolutional operations (i) helps to reduce the computational load by shifting the architecture of 5-by-5 convolutions in one-layer into 5-by-1, and 1-by-5 convolutions in two layers [26], and (ii) modulates the receptive field through the application of dilated convolutions. Finally, the proposed model builds on the basic computational unit to fuse region and boundary-based information. The boundary-based model helps to complete perceptual boundaries, which is important for delineating overlapping nuclei.

III. METHOD

In this section, we summarize steps for improving nuclear segmentation through region-based, a fusion of region and boundary-based information, and post-processing.

*A. Deep Learning Models for Region-based Segmentation*

Traditionally, convolutional neural networks (CNN)s have been used to perform image classification in the computer vision literature. CNN consists of several layers of convolution operation, where each convolutional layer is usually followed by the max-pooling. The last layer is a fully connected layer, which maps a high dimensional vector to a low dimensional probability vector corresponding to distinct classes. In recent years, a diversity of CNN architectures have been proposed based on the depth and size of the model for the classification task (e.g., ImageNet [1], VGG [2], ResNet [27]). The segmentation task can also be performed by using a sliding window coupled with the classification to label each pixel in the image. However, this approach has been shown to be either noisy, or less accurate, or time-consuming. To overcome these issues, alternative CNN architectures (e.g., FCN[6], UNet[5], SegNet[3], ENet[4]) have been proposed for region-based segmentation. These models are based on the encoder-decoder architecture. The encoder architecture is the same as vanilla CNN, which consists of several convolution layers followed by max-pooling. The encoder layers perform feature extraction and region-based classification of the down-sampled image. On the other hand, the decoder layers perform up-sampling after each convolutional layer, to compensate the down-sampling effects of the encoder, and, to generate an output with the same size as the input. Some of these models are symmetric (e.g., the encoder and decoder have the same depth) and some are asymmetric. In the latter case, the decoder has the advantage of the smaller number of convolutional layers for reducing the computational load. One of the limitations of the encoder-decoder architecture is that up-sampling in the decoder layers may not compensate for the drastic down-sampling that takes place in the encoder layers. As a result, detailed spatial information may be lost. Several strategies have been proposed to overcome the loss of detailed information. For example, (i) FCN and UNet addressed this problem by feed-forwarding the output of the feature maps, from the encoder layers to the matched decoder layers, and (ii) ENet and SegNet save the indices of the selected pixels in the max-pooling layer, which are integrated sparsely in the up-sampled output in the decoder layers.

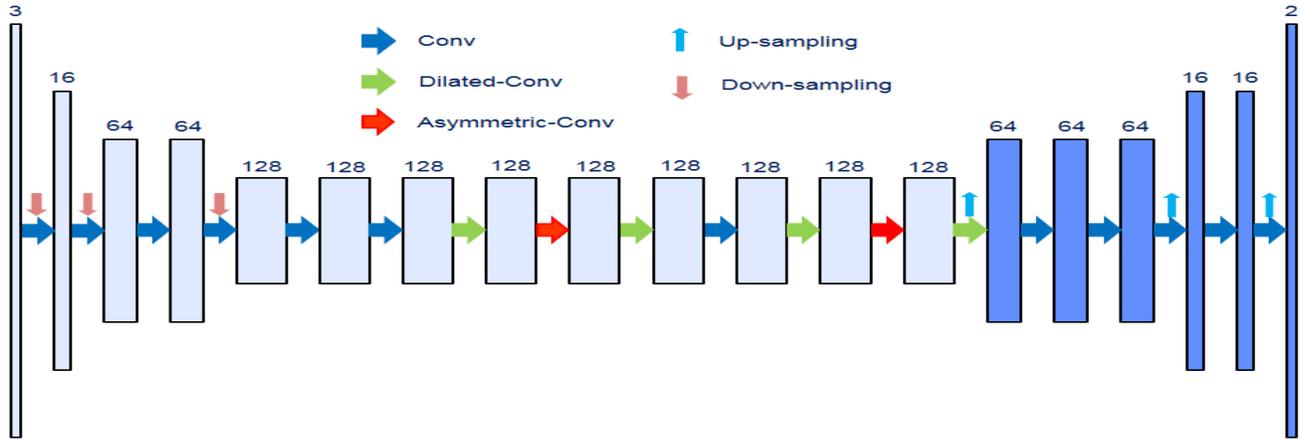

Fig. 2. The complete architecture of the ENet model for nuclear segmentation is shown in terms of layers of convolutional networks. The model includes both encoder (light blue) and decoder (dark blue) parts. The upward and downward arrows indicate up-sampling and down-sampling operations. Right hand arrows show different types of convolution including normal, dilated, and asymmetric.

*B. Region- and Boundary-based Segmentation with ENet*

We experimented with several configurations of the ENet model to assess the performance of the nuclear segmentation. ENet has been shown to have a superior segmentation performance when compared to alternative encoder-decoder architectures (e.g., SegNet) [4]. First, the performance of the segmentation was evaluated with region-based models. Next, we examined if the performance can be further increased by fusing of region- and boundary-based representations. These processes are preceded by color decomposition (CD) step. CD reduces the required sample size, for training, and speeds up the computational throughput. This step decomposes the RGB signal into two channels of information corresponding to the DNA and protein contents, where the former channel is used for subsequent processing. CD is based on a recently published method that has been shown to provide superior results [28].

ENet is an asymmetric encoder-decoder model for real-time semantic segmentation. It consists of 17 layers where the initial layer performs subsampling to reduce the computational load. The remainders are (i) 10 convolutional layers, in parallel, with max-pooling for the encoder, (ii) 5 convolutional layers in parallel, with up-sampling, for the decoder, and (iii) a final 1×1 convolutional layer to merge the output bank of the penultimate layer. Figure 2 shows the ENet architecture. All the convolution operations are either 3-by-3 or 5-by-5. However, 5-by-5 convolutions are asymmetric, i.e., they are implemented separately as 5-by-1 and 1-by-5 convolutions to reduce the computational load. In addition, some of the layers have been used dilated convolution to increase the effective receptive field of the associated layer. This helps with the faster growth of the receptive field of the encoder without using down-sampling. The ENet model is highly efficient since all convolutions are either 3-by-3 or by 5-5 and parallel, as opposed to sequential, integration with max-pooling potentially preserves intrinsic details of the phenotypic signature.

Having summarized the ENet architecture, we train with annotated regions and boundaries. The latter is computed from annotated regions corresponding to each nucleus. Training for regions has the advantage of learning a model for a diversity of phenotypes (e.g., vesicular, hyperchromatic). On the other hand, training for boundaries has the advantage of (i) overcoming fuzzy boundaries of the nuclei predicted by region-based ENet, (ii) compensating the loss of detailed spatial information as a result of max-pooling, (iii) separating touching or overlapping nuclei with perceptual boundaries. Both region- and boundary-based segmentation utilize the ENet architecture. The only difference between the boundary- and region-based networks is how labels are assigned to each pixel. Finally, fusion of the region- and boundary-based ENets is achieved by a tertiary ENet that utilizes the output of region- and boundary-based segmentation results. We refer to this architecture as the Fused-ENet.

*C. Post-processing*

Fused ENet improves the segmentation results in a number of ways that includes separation of touch nuclei; however, not all of the overlapping nuclei are delineated. Therefore, a final post-processing step is added. Post-processing methods for separating touching nuclei include but not limited to marker-based watershed [29], gradient flow tracking [30], regularized centroid transform [31], and geometric reasoning based on points of maximum curvature [32]. The marker-based watershed method computes a marker from the minima computed from the distance transform of the binarized mask and then applies the watershed method. Gradient flow tracking and regularized centroid transform are based on modeling delineation of touching nuclei based on partial differential equations. Finally, geometric reasoning, based on points of maximum curvature, partitions the space and generates a set of hypotheses for decomposing a clumped region. Everyone of these methods has their advantages and disadvantages. We used marker-based watershed method because of its computational simplicity and open source availability.

## IV. EXPERIMENTS AND RESULTS

In this section, we (i) describe the training and testing dataset and the network parameters setting, and (iii) evaluate the performance of the proposed model against previous methods.

### A. Experimental Set-up

In order to capture technical and biological variations, 32 images consisting of 21 brain of 11 breast histology sections were hand-annotated, which resulted in approximately 19,000 nuclei. All sections are imaged with a 20X objective with a pixel size of approximately 0.5 microns. Each image is sampled to generate 96 non-overlapping patches of the size 360-by-480 for training and testing. This augmented dataset includes alternative phenotypes of the nuclear signature (e.g., normal, vesicular, hyperchromatic), which is approximately divided between the training and testing cohorts (e.g., 50-50). The training procedure consists of two steps. (I) The region- and boundary-based models are trained separately. (II) The outputs of the region- and boundary-based networks were used as the input to train the fusion network. The same encoder-decoder architecture was used for all the three networks (i.e., region-based, boundary-based, and fusion networks). The Adam optimization algorithm [33] has been used for training, where the batch size is set at four due to the limitations of the GPU memory. The learning rate and L2 weight decay are set at 5e-4 and 2e-4, respectively. The dropout method is used to avoid overfitting.

### B. Evaluation

Fused-ENet with postprocessing produces the best segmentation result, as shown in Table 1, where the performance of ENet is also evaluated against the multi-reference graph cut method [13]. Improved performance is the direct results of simultaneous segmentation of nuclei morphometry with diverse signatures. In addition, color

TABLE I
COMPARISON FOR NUCLEAR SEGMENTATION BETWEEN ENET (WITHOUT POSTPROCESSING) AND MULTIREFERENCE GRAPH CUT

| Approach | Precision | Recall | F-Score |
|---|---|---|---|
| Fused-ENet+CD+Watershed | **0.94** | **0.88** | **0.91** |
| Fused-ENet+CD | 0.91 | 0.87 | 0.89 |
| Fused-ENet+RGB | 0.82 | 0.87 | 0.84 |
| Single ENet+CD+Watershed | 0.90 | 0.86 | 0.88 |
| Single ENet+CD | 0.84 | 0.87 | 0.85 |
| Single ENet+RGB+Watershed | 0.84 | 0.86 | 0.85 |
| Single ENet_RGB | 0.81 | 0.78 | 0.79 |
| Multi-reference Graphcut | 0.75 | 0.85 | 0.79 |

decomposition enhances the performance further. Our experiments indicated that:

(i) ENet enables segmentation of complex nuclear phenotypes, such as vesicular phenotype, which has been complex to model using traditional computer vision methods. **Figure 3** illustrates segmentation results with diverse nuclear phenotypes.

(ii) Boundary-based information helps to separate touching or overlapping nuclei with perceptual boundaries. Delineating of overlapping nuclei is one of the barriers in nuclear segmentation for single cell profiling. In our analysis, we selected 16 test images, where 98 touching nuclei were randomly selected. The analysis indicates that 62 touching nuclei (e.g., 63.2% improvement) are correctly separated by fused ENet. The behavior of the fused ENet for delineating touching nuclei is shown for two examples in **Figure 4.**

(iii) Finally, the proposed model is time efficient and capable of performing instant-based segmentation. The fusion network has been implemented on a server with a single GPU card. Each image is of the order of 1k-by-1k, and, on the average, the processing time for an image of the size 360-by-480 is approximately 60 ms.

## V. CONCLUSION

There are two intrinsic barriers for nuclear segmentation in H&E stained images. These are (i) the diversity of phenotypes and (ii) overlapping nuclei that form perceptual boundaries. We have demonstrated that these two issues can be largely resolved with the fusion of ENets. In contrast, image-based modeling for segmentation of alternative phenotypic signatures has been a challenge using the traditional image analysis methods. Fusion of ENets is based on integrating learned networks from region- and boundary-based representations. The overall framework consists of color decomposition, fusion of ENets, and a postprocessing step. Color decomposition generates a single relevant gray scale image corresponding to the nuclear dye; hence, reducing the required number of training samples.
Nuclear segmentation is based on the ENet architecture with the encoder-decoder layers. The region- and boundary-ENets are trained to delineate diverse spatial signatures and their corresponding boundaries. A third ENet is trained to combine the output of region- and boundary-based networks. We have shown that nuclei with diverse phenotypic profiles can be delineated, and, in the majority of cases, overlapping nuclei can be partitioned. The remainder of the overlapping nuclei is partitioned with a post-processing step. Intuitively, CNN learns complex spatial signature and can fill-in missing boundaries. Segmentation is fast and enables rapid delineating of a large number of samples for high throughput processing.

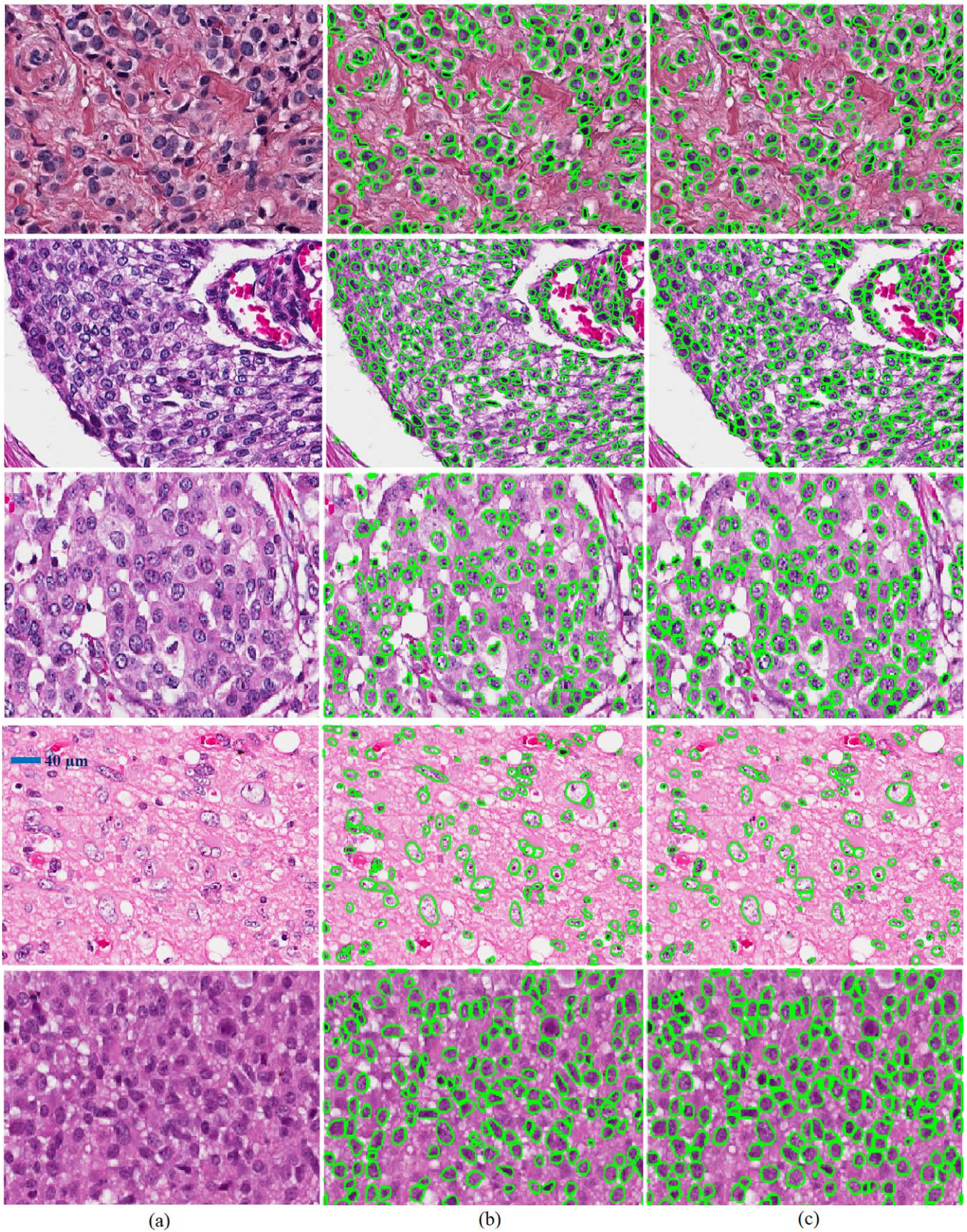

*Fig. 3.* Qualitative performance of nuclear segmentation is shown for five different phenotypes. Columns (a), (b), and (c) illustrate the original image, ground truth, and segmentation results of the proposed model, respectively. Rows 1-3, and 4-5 correspond to breast and brain tumor sections, respectively. These images show large variations in color and phenotypic signatures. More specifically, rows 2-3 show nuclear phenotypes with heterogeneous hematoxylin stain; and row 4 shows nuclei with vesicular phenotypes.

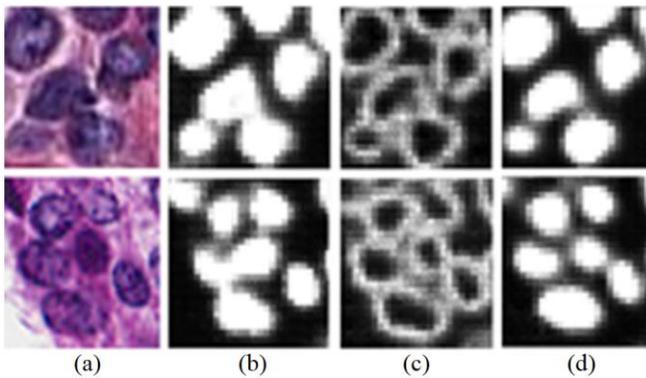

Fig. 4. Integration of region-based and boundary-based segmentation helps to separate touching nuclei. (a) shows a few examples of touching and overlapping nuclei, (b) and (c) are the output probability maps of the region-based and boundary-based segmentation models, and (d) indicates the output probability map of the fusion model.

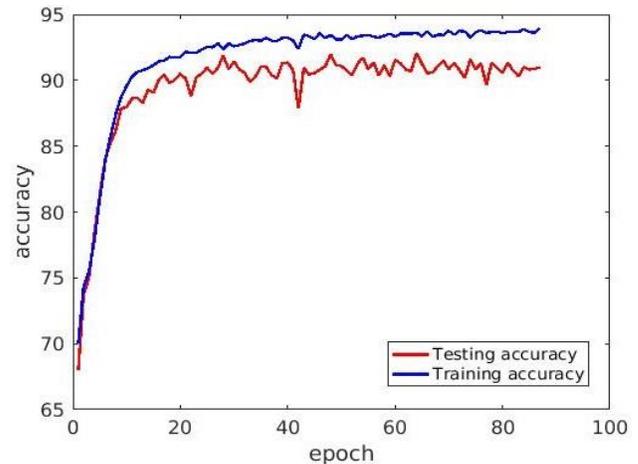

Fig.5. Testing and training accuracies are increased as a function of number of iterations.